\documentclass[lettersize,journal]{IEEEtran}

\usepackage{amsmath,amsfonts}
\usepackage{algorithm}
\usepackage{array}
\usepackage[caption=false,font=normalsize,labelfont=sf,textfont=sf]{subfig}
\usepackage{textcomp}
\usepackage{stfloats}
\usepackage{url}
\usepackage{verbatim}
\usepackage{graphicx}
\usepackage{cite}

\usepackage{threeparttable}

\usepackage{float}  

\usepackage{algpseudocode}

\begin{document}

\title{Robust Visual Teach-and-Repeat Navigation with Flexible Topo-metric Graph Map Representation}

\author{Jikai Wang, Yunqi Cheng, Kezhi Wang, and Zonghai Chen, ~\IEEEmembership{Senior Member,~IEEE,}
\thanks{This work is supported by the National Natural Science
	Foundation of China (Grant No. 62103393). (Corresponding author: Zonghai Chen)}
\thanks{Jikai Wang, Yunqi Cheng, Kezhi Wang, and Zonghai Chen are with Department of Automation, University of Science and Technology of China (USTC), Hefei, 230027, PR China (e-mail: wangjk@ustc.edu.cn; chengyunqi@mail.ustc.edu.cn; wangkz@mail.ustc.edu.cn; chenzh@ustc.edu.cn).
	}%
}

\markboth{}%
{Shell \MakeLowercase{\textit{et al.}}:Robust Visual Teach-and-Repeat Navigation with Flexible Topo-metric Graph Map Representation}

\maketitle

\begin{abstract}
Visual Teach-and-Repeat Navigation is a direct solution for mobile robot to be deployed in unknown environments. However, robust trajectory repeat navigation still remains challenged due to environmental changing and dynamic objects. In this paper, we propose a novel visual teach-and-repeat navigation system, which consists of a flexible map representation, robust map matching and a map-less local navigation module. During the teaching process, the recorded keyframes are formulated as a topo-metric graph and each node can be further extended to save new observations. Such representation also alleviates the requirement of globally consistent mapping. To enhance the place recognition performance during repeating process, instead of using frame-to-frame matching, we firstly implement keyframe clustering to aggregate similar connected keyframes into local map and perform place recognition based on visual frame-to-local map matching strategy. To promote the local goal persistent tracking performance, a long-term goal management algorithm is constructed, which can avoid the robot getting lost due to environmental changes or obstacle occlusion. To achieve the goal without map, a local trajectory-control candidate optimization algorithm is proposed. Extensively experiments are conducted on our mobile platform. The results demonstrate that our system is superior to the baselines in terms of robustness and effectiveness. 
\end{abstract}


\section{Introduction}
\label{sec:introduction}
\IEEEPARstart{R}{ecently}, mobile robots are widely applied in  industrial and household scenes \cite{10036102}. Visual localization \cite{10521763,8931657} and navigation \cite{9316980} methods are extensively studied. Under
the condition that task route is certain, such as navigating between fixed stations, Visual Teach-and-Repeat (VTR) navigation \cite{van2024visual} can avoid fully mapping of the task environment and make deploying robot efficiently. The teaching process is generally controlled by human operator and the robot records visual frames as map along the task route in real-time. During the repeating process, the robot tries to match the current visual frame with the map and update its next-step goal if matched successfully. Thus, to achieve robust navigation, it is critical for robot to implement correct place recognition frequently and keep tracking its goals, especially in conditions of environment changing, dynamic object occlusion, and route deviation. 

Map in the existing VTR methods \cite{10433735,sun2021robust,dall2021fast} is formulated as a set of keyframes collected during the teaching process, in which the keyframes are connected spatially and temporally. Thus, once the robot matches one keyframe in map, it can directly derive the next keyframe’s pose as goal and navigate to it. Ideally, such process is running iteratively until it meets the ending keyframe. Frame retrieval and matching are also termed as place recognition \cite{8931657}, which has been extensively studied for decades. Considering the time-efficiency, traditional place recognition methods are applied, including directly image comparison \cite{dall2021fast,van2024visual}  and feature-based image matching \cite{furgale2010visual,paton2016bridging}. Directly image comparison methods firstly compress the image and then compute the similarity between compressed images. It is efficient but less robust to viewpoint changes and occlusion. To alleviate the problem, feature-based image matching methods extract sparse features on images and implements place recognition using DBoW library. With 3D information, they can also compute relative transformation using PnP solver \cite{zheng2013revisiting}. It achieves balance between efficiency and robustness. Recently, with the development of artificial intelligence, powerful deep learning-based image matching models are also incorporated into VTR systems \cite{swedish2018deep,10433735,gao2017intention,ai2022deep,sorokin2022learning}. Though the robustness is achieved, time-computing cost is increased. Furthermore, deep learning-based image matching models’ limited interpretability makes it questionable to deploy robots in novel environments. Actually, in practical application, environmental changes and dynamic objects are common, which bring great challenges. The existing place recognition methods that can be used on VTR robots cannot fully handle these challenges. A robust navigation module is highly required. 

Promoting local navigating ability is another solution to compensate place recognition deficiency.  The aim of navigation module is to continuously track goals given limited environmental heuristic information provided by place recognition module. How to recover trajectory under the conditions of dynamic avoidance, occlusion,and key place recognition fail is critical issue. The traditional navigation systems \cite{ullah2024mobile,zhou2020ego}  are highly dependent on accurately localization system. However, globally accurately localization is avoided in VTR navigation system. Thus, in the existing VTR systems, the navigation actions are directly encoded in the map. Perception module derives prior navigation actions from the matched content in the map and implement correction based on matching degree. Such mechanism is fragile when the matching algorithm fails. 

In this paper, we propose a novel VTR navigation system and its performance is promoted from two aspects. Firstly, we propose a flexible map representation, in which globally topology information and locally metric information are incorporated as a keyframe graph. In our graph, to handle the environmental change issue, each vertex is extendable and novel observation collecting during repeating process at same area are added to the vertex. To reduce the map redundancy, we perform keyframe clustering and only preserve representative keyframes, which saves neighboring redundant keyframe’s information. Secondly, we propose a robust navigation module, in which we manage  relative goal list and iteratively select the best local trajectory candidate to approximate the goal list. The goal list is update once the place recognition is success.   Each local trajectory candidate corresponds to preset control commands, which makes the local navigation easy to use.

Our contributions are as follows. 

\begin{enumerate}
	\item A flexible map representation is proposed, which can be adapted to environmental changes and drift errors of odometry. 
	\item A robust visual teach-and-repeat navigation system is proposed, which has been proved to be adapted to environmental changes, dynamic objects, and viewpoint occlusion. The navigation module can also be embedded into other VTR systems. 
	\item Our system is friendly to users and can be adapted to new task environment easily.
\end{enumerate} 

The rest of this paper is organized as follows. In Section II,
the related work is presented. In Section III, the proposed method is detailed.   Section IV presents the experimental results and analysis. Finally, Section V concludes this paper.

\section{Related Work}

It is widely believed that obtaining and maintaining detailed metric environment information are challenged and time-consuming, especially in large-scale or dynamically changing environments. Thus, since map is used for navigation, it is more convenient to directly combing local scene cues and navigation cues together, which can make the robot directly obtain navigation heuristic information through local scene recognition. Such mechanisms are deeply inspired by human beings. In SAM \cite{loo2024scene}, a concept of Scene Action Maps (SAMs) is proposed, which represents the environment as a graph of interconnected navigational behaviors. The SAMs enable robots to navigate effectively even with limited metric and spatial information. In\cite{cuizhu2023one}, the teaching image sequence is divided according to navigation actions. During the repeating phase, a Siamese network is applied for image retrieval and the results are directed to navigation actions to implement. Deep learning-based methods are widely used for image matching, which are robust to environmental and viewpoint changes. Instead of representing the environments as a topological graph of visual frames, Garg et al. \cite{garg2024robohop}  construct graph of segments and establish a novel mechanism for intra- and inter-image connectivity based on segment-level descriptors and pixel centroids. Their method are more friendly to object navigation. In \cite{10433735}, critical navigation waypoints are extracted and the mapping between waypoints visual cues and navigation cues are learned. During the teaching phase, the agent frequently tries to recognize waypoints. \cite{furgale2010visual} is the most classical VTR framework, which constructs overlapping submaps motivated by drift errors of visual odometry. Then, the robots continuously perform submap localization instead of global localization. Path tracking is achieved based on a ground plane assumption.  \cite{paul2024mpvo} also argues that Visual Odometry (VO) is not globally accurate thus cannot be directly used for point goal navigation. To alleviate this problem, they fuse a traditional visual odometry for coarse pose estimation and a learning-based visual odometry for fine pose correction. Once the VO pose is updated, the navigation information is also updated accordingly.  \cite{van2024visual} is  constructed based on the viewpoint that visual odometry is with drift errors and not be able to support global navigation. Thus, they combine VO and periodic orientation correction using visual comparison, which ensure that the agent can follow back its route. The robot uses odometry to move towards the location of the next snapshot, and then employs visual homing to converge to the snapshot location, ensuring that it stays within the catchment area. This combination allows the robot to navigate efficiently without the need for detailed maps or complex processing. 
\cite{10578334} introduces a predictive approach to data acquisition, where the robot uses its past experiences and current observations to make informed decisions about which data to collect. This approach leads to a more balanced and up-to-date training dataset, which is crucial for achieving robust robot operation. 
\cite{rozsypalek2023multidimensional} considers that the changing environment makes it ambiguity  to recognize places. They use particle filter to simultaneously estimate the frame correspondence of current observation on teaching map. The particles' weights are updated according to a image similarity evaluation neural network.  With such estimation, they can effectively correct the moving command to make robot approximates the teaching route. In \cite{krajnik2018navigation}, a global localization-free simple visual teach-and-repeat navigation system is constructed, in which the robot only correct its heading based on frame matching results during the repeating phase. Though they proved that  a robot can use its camera information only to
correct its heading and it does not have to build metric maps
or perform explicit localization, heavily dependent on the frame matching results makes it difficult for robot to handle obstacle occlusion situations.

Fusing topological and metric information has been a common solution for global metric map-less-based navigation. However, in these methods, though various kinds of topological maps are proposed,  how to adapt to changing and dynamic environment for practical robot deployment remains challenging.

\section{Proposed Method}

\subsection{System Framework}

\begin{figure*}[t]
	\centerline{\includegraphics[width=1.9\columnwidth]{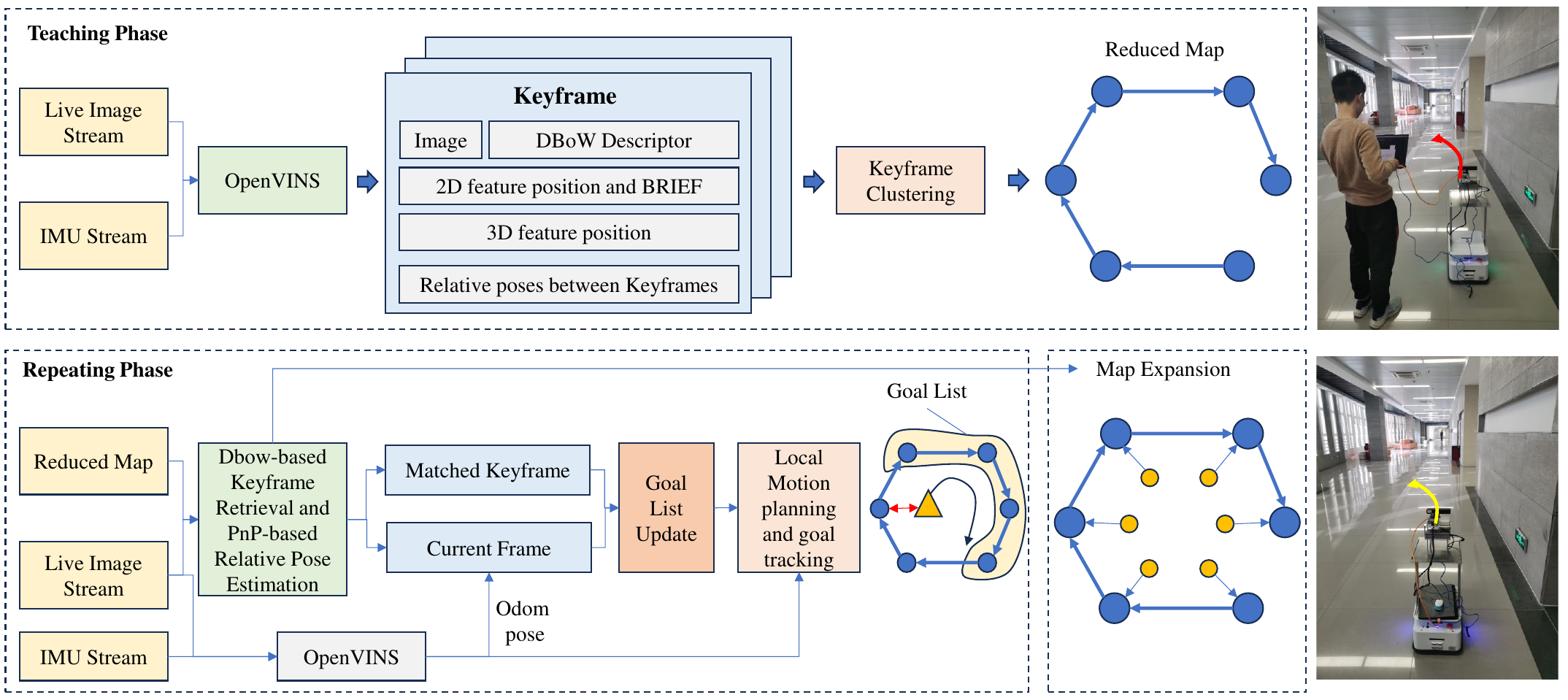}}
	\caption{Our system flowchart.}
	\label{systemframework}
	\vspace{-0.5cm}
\end{figure*}

Our system is consisting of two phases. In the teaching phase, the robot is controlled by human operator. In the meantime, Openvins \cite{geneva2020openvins}  is running as visual odometry and a loop module is activated to construct a topological keyframe graph. Notice that we do not perform globally graph optimization. Our system is acceptable with drift errors and focuses on the accuracy of relative pose transformation between connected keyframes. After the teaching process is finished, the map redundancy reducing step is performed. In the repeating phase, the robot periodically performs loop detection and manages the goal list according to the loop results. Then, the robot moves to approximate the goal list with our motion planning module.  The system is illustrated in Fig. \ref{systemframework}.

\subsection{Visual Teaching}

\subsubsection{Map Representation Error Analysis}

Map representation is critical issue for mobile robot navigation. Traditionally, visual metric map is represented by estimated trajectory and its associated observations. Thus, the map representation error heavily relies on mapping trajectory estimation precision. In the existing visual SLAM methods \cite{9440682,qin2017vins,9484819,10042487}, the map is generally represented as
\begin{equation}
	\hat{\mathcal{M}} = \{[\mathcal{K}_i,\hat{\mathbf{T}}^W_i],i=1,\cdots,N\},
\end{equation}
where $\mathcal{K}_i$ denote keyframe containing visual observations and $\mathbf{T}^W_i$ is its estimated global pose. With loop correction, $\mathbf{T}^W_i$ is computed as
\begin{equation}
	\hat{\mathbf{T}}^W_i = \mathbf{T}^W_0\mathbf{T}^0_1\triangle\mathbf{T}^0_1\cdots\mathbf{T}^{i-1}_i\triangle\mathbf{T}^{i-1}_i,
\end{equation}
where $\mathbf{T}^0_1\triangle\mathbf{T}^0_1$ is the estimated relative pose transformation, $\mathbf{T}^0_1$ is the unknown ground truth, and $\triangle\mathbf{T}^0_1$ is drift pose. Thus, the ground truth should be 
\begin{equation}
	\mathbf{T}^W_i = \mathbf{T}^W_0\mathbf{T}^0_1\cdots\mathbf{T}^{i-1}_i,
\end{equation}
which is impractical to compute in SLAM paradigm. The difference between $\hat{\mathbf{T}}^W_i$ and $\mathbf{T}^W_i$ makes the map spatially distorted. Drift errors are accumulated, which makes it infeasible to implement robust navigation according to $\hat{\mathcal{M}}$. Notice that the key issue is due to that each keyframe pose is assumed to be globally aligned. Thus, for the sake of robust navigation, a more general map representation should be 
\begin{equation}
	\bar{\mathcal{M}} = \{[\mathcal{K}_i,\hat{\mathbf{T}}^i_j],i,j=1,\cdots,N\}.
\end{equation}
Each keyframe only preserves its credible relative pose transformation with its neighboring keyframe. It is based on the condition that relative pose estimation now is significantly accurate. Standing at any keyframe, we can be directed to connected keyframe iteratively untill that all the keyframes are traversed. If we want to get relative pose transformation between two unconnected keyframe, we can only accumulate the credible relative pose transformations between them. It is also easy to update the map locally. Thus, using such representation, the key issues for robust navigation are recognizing the cloest keyframe and constructing credible relative pose transformation. 

\subsubsection{Topo-metric Keyframe Map}
When the robot is running, its mounted visual camera continuously obtains frame stream. The Openvins module estimate the pose, extract corner features, and recover 3D feature positions of each frame. We extract keyframes every 5 frames and a keyframe is denoted as
\begin{equation}
	\mathcal{K}_i = \{\mathbf{T}^{i-1}_i,U_i,P_i, \mathbf{I}_i\},
\end{equation}
where $\mathbf{T}^{i-1}_i$ is the relative transformation between $\mathcal{K}_{i-1}$ and $\mathcal{K}_i$. $U_i=\{[u_n,v_n,\mathbf{d}_n],n=1,\cdots,N\}$ is the 2D feature point pixel coordinates set and $\mathbf{d}_n$ denotes its descriptor.  $P_i=\{\mathbf{p}_n = [x_n,y_n,z_n]^T,n=1,\cdots,N\}$ is their 3D positions set in camera coordinates. $\mathbf{I}_i$ is the image. During the map extension with new keyframe coming, loop detection module is activated and tries to detect loop between the current keyframe with   historical keyframes using DBoW2 library. Once a loop detected, the relative transformation between current keyframe $\mathcal{K}_i$ and its looped keyframe $\mathcal{K}_{L(i)}$ is computed, and $\mathcal{K}_i$  is updated as
\begin{equation}
	\mathcal{K}_i = \{\mathbf{T}^{i-1}_i,U_i,P_i, \mathbf{I}_i,\mathbf{T}^{L(i)}_i,L(i)\},
\end{equation}
where $L(i)$ is the index of looped keyframe.

The map is consisting of keyframes and each keyframe record its temporally connected keyframes. Thus, globally consistent is not highly required in the map and thus alleviate the requirement of high-performance SLAM systems.
\subsubsection{Map Redundancy Reduction }
\begin{figure}[t]
	\centerline{\includegraphics[width=0.95\columnwidth]{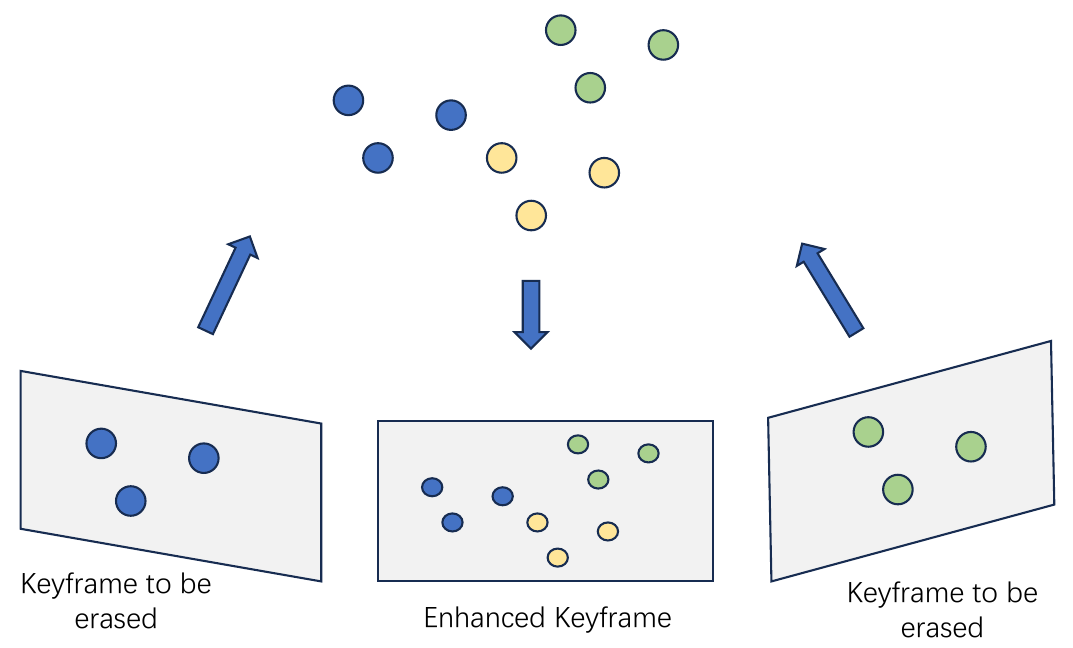}}
	\caption{Our keyframe clustering illustration. We cluster keyframes that are similar measured by DBoW2 tools. The keyframes to be erased project its feature points into the keyframe to be preserved. The preserved keyframe thus contains novel information from neighboring keyframes. }
	\label{keyframecluster}
\end{figure}

Even though we have applied keyframe mechanims, our map is still redundant and its size increases significantly with the teaching route length increasing. Furthermore, densely keyframes also make performing loop detection frequently, which is a time-consuming process. Thus, we perform map redundancy reduction. For the $i$-th keyframe, we compute its DBoW similarity with following keyframes until the similarity is below a threshold and stop at $\mathcal{K}_{S(i)}$. Then, the similar keyframes are merged into $\mathcal{K}_i$, which is expanded as
\begin{equation}
	\mathcal{K}_i =  \{\mathbf{T}^{i-1}_i,\bar{U}_i,\bar{P}_i, \mathbf{I}_i,\mathbf{T}^{L(i)}_i,L(i),\mathbf{T}^{i}_{S(i)},S(i)\},
\end{equation}
Specifically, for one similar keyframe $\mathcal{K}_{i+s}$, we transform its 3D feature positions into the coordinates of $\mathcal{K}_i$, and then add them into $P_i$. The final extended $P_i$ is denoted as $\bar{P}_i$. The image projection of $\bar{P}_i$ is denoted as $\bar{U}_i$. $\mathbf{T}^{i}_{S(i)}$ is the transformation between  $\mathcal{K}_{S(i)}$ and $\mathcal{K}_i$. Then, the keyframes between  $\mathcal{K}_i$ and $\mathcal{K}_{S(i)}$ are directly removed in the map. The map is essentially a link list. Given a keyframe, we can always find its next connected keyframes.  An illustration is presented in Fig. \ref{keyframecluster}. Essentially, in the raw map, due to that the used FAST feature points are sensitive to viewpoint changes, we have to preserve redundant keyframes. In our method, to reduce redundancy, we cluster keyframes and aggregate feature points into sparse keyframes.

\subsection{Visual Repeating}

One of the critical issues in visual repeating is to find loop matches of current frame $\mathbf{I}_k$ with keyframes in map. In this paper, we detect loop match of $\mathbf{I}_k$ using DBoW2\footnote{https://github.com/dorian3d/DBoW2}. If detected successfully with $\mathcal{K}_i$, the pose of $\mathcal{K}_{S(i)}$ is defined as current goal.  Since local navigation is geometry-aware, it is necessary to compute the accurate goal position in odometry coordinates. Thus, computing the relative pose transformation between the current frame with its matched keyframe is required. 

\subsubsection{Frame-to-Keyframe Matching}
Notice that information in local area has been aggregated in $\mathcal{K}_i$. Given $\bar{U}_i$, $\bar{P}_i$, and $\mathbf{I}_k$, the matching process is detailed as follows. For each feature $\mathbf{f}_n = [u_n,v_n,\mathbf{d}_n]$ in $\bar{U}_i$, we aim to find its corresponding feature in $\mathbf{I}_k$. To avoid ambiguity, we only extract features on the region of  $\mathbf{I}_k$ defined as
\begin{equation}
	R_n = \{[u,v]^T|\|[u,v]^T-[u_n,v_n]^T\|_2<\gamma\},
\end{equation}

It means we only find potential corresponding features on $\mathbf{I}_k$ for $\mathbf{f}_n$  within a circular area centered at $[u_n,v_n]$. It is based on the assumption that  $\mathbf{I}_k$ and $\mathcal{K}_i$ have similar viewpoints. We extract FAST corner features within $R_n$ and compute its BRIEF descriptor. Then, we compute the descriptor distance between  $\mathbf{d}_n$ and the extracted features' descriptors. If the minimum distance is below a threshold, then the best feature match of $\mathbf{d}_n$ is found and denoted as $\mathbf{f}^k_n = [u^k_n,v^k_n,\mathbf{d}^k_n]$. After processing all the features, we have a 3D-2D correspondence set denoted as
$\left\{\mathcal{C}_n = \{\mathbf{p}_n,[u^k_n,v^k_n]^T\},n=1,\cdots,N\right\}$. Then, we aim to compute the optimal relative pose $\mathbf{T}_k^i=[\mathbf{R}_k^i,\mathbf{t}_k^i]$ of $\mathbf{I}_k$, which satisfies 
\begin{equation}
	\min\sum_{n=1}^N\|\left[\mathbf{K}\cdot(\mathbf{R}_k^i\cdot \mathbf{p}_n + \mathbf{t}_k^i)\right]_{1:2}-[u^k_n,v^k_n]^T\|^2_2,
\end{equation}
where $\mathbf{K}$ is the intrinsic matrix. We implement pose optimization using Gauss-Newton (GN) algorithm. We denote $\mathbf{R}_k^i\cdot \mathbf{p}_n+ \mathbf{t}_k^i $ as $\mathbf{p}_n^{\prime}$. Each cost term is defined as
$\mathbf{e}_n$, and the Jacobian matrix with respect to the pose update is computed as
\begin{equation}
	\frac{\partial \mathbf{e}_n}{\partial \delta\mathbf{\xi}} = \frac{\partial \mathbf{e}_n}{\partial \mathbf{p}_n^{\prime}}\cdot\frac{\partial \mathbf{p}_n^{\prime}}{\partial \delta\xi},
\end{equation}
where we have 
\begin{equation}
	\frac{\partial \mathbf{e}_n}{\partial \mathbf{p}_n^{\prime}} = \begin{bmatrix}
		\frac{f_x}{z^{\prime}}     & 0 & -\frac{f_xx^{\prime}}{z^{\prime2}}     \\
		0 & \frac{f_y}{z^{\prime}} & -\frac{f_yy^{\prime}}{z^{\prime2}}  \\
	\end{bmatrix},
\end{equation}
and 
\begin{equation}
	\frac{\partial \mathbf{p}_n^{\prime}}{\partial \delta \xi} = [I,-\mathbf{p}^{\prime\land}].
\end{equation}

By fusing all the error terms, according to GN, we have
\begin{equation}
	\delta \xi = -\left(\mathbf{J}^T\mathbf{J}\right)^{-1}\mathbf{J}^T\mathbf{e},
\end{equation}
where $\mathbf{e}$ is the stacked error terms and $\mathbf{J}$ is the stacked Jacobian matrix. The pose estimation is initialized as identical matrix. Such process is implemented iteratively until meeting convergence. If it cannot achieve convergence, it means that $\mathcal{K}_i$ and $\mathbf{I}_k$ are not good match, which will be directly abandoned. Otherwise, the obtained estimation of $\mathbf{T}_k^i$ is used to update the map and goal list.

Furthermore, it should be noticed that the optimization problem tends to be biased and degraded if the matched features concentrate on local areas densely instead of distributing on the image evenly, which is common in practical environment. Also, repetitive features tend to exist due to the former keyframe clustering step. To alleviate such problem, we apply grid filter on the raw correspondences and try to make them evenly distributed. 

\subsubsection{Map Expansion}
Environmental changing is critical issue for VTR navigation. It requires that the map should timely keep consistent with the real environment. To alleviate this problem, we also expand the map during visual repeating process. Ideally, if the robot strictly follows the teaching route, every frame should be matched with one keyframe. However, in practical application, the condition cannot be satisfied due to environmental changes and dynamic objects. It means that robot collects novel information during repeating process and we aim to add the  novel information  in the map. Specifically, map expansion is activated when the robot every secondly matches successfully with keyframes.  Under the condition of $\mathbf{I}_k$ being matched with $\mathcal{K}_i$ and $\mathbf{I}_{k+t}$ being matched with $\mathcal{K}_j$, then each frame $\mathbf{I}$ between $\mathbf{I}_k$ and $\mathbf{I}_{k+t}$ is added to segment between $\mathcal{K}_i$ and $\mathcal{K}_j$. Thanks to the locally accurate relative pose estimation of odometry, $\mathbf{I}$ can find its nearest keyframe, then it will be add to the keyframe, which is denoted as
\begin{equation}
	\mathcal{K}_i =  \{\mathbf{T}^{i-1}_i,\bar{U}_i,\bar{P}_i, \mathbf{I}_i,\mathbf{T}^{L(i)}_i,L(i),\mathbf{T}^{i}_{S(i)},S(i),\{\mathcal{K}\}\},
\end{equation}
where $\{\mathcal{K}\}$ denotes the expanded frames recording novel information of the local area. Then, in the following loop detection process, the expanded frames will also be involved. To avoid destroy the map structure, we do not set the expanded frames as new keyframes. We set them as  low-level keyframes attached to map keyframes. Such mechanism makes the map expansion process do not affect the goal management. We can still obtain goals sequentially. 

\subsubsection{Goal List Management}
Managing only one goal in the existing VTR methods makes the navigation system fragile to dynamic objects, occlusion, and fast movement, which make the goal unreachable or tracking lost. In this paper, we manage a goal list to ensure the global consistency of VTR navigation. Specifically, for frame $\mathbf{I}_k$ and its successfully matched keyframe $\mathcal{K}_i$, the current goal pose $\mathbf{T}_g^k$ with respect to the current robot coordinates is computed as
\begin{equation}
	\mathbf{T}_{g0}^k = inv(\mathbf{T}_k^{i})\cdot \mathbf{T}^i_{S(i)}.
\end{equation}

\begin{figure}[t]
	\centerline{\includegraphics[width=0.95\columnwidth]{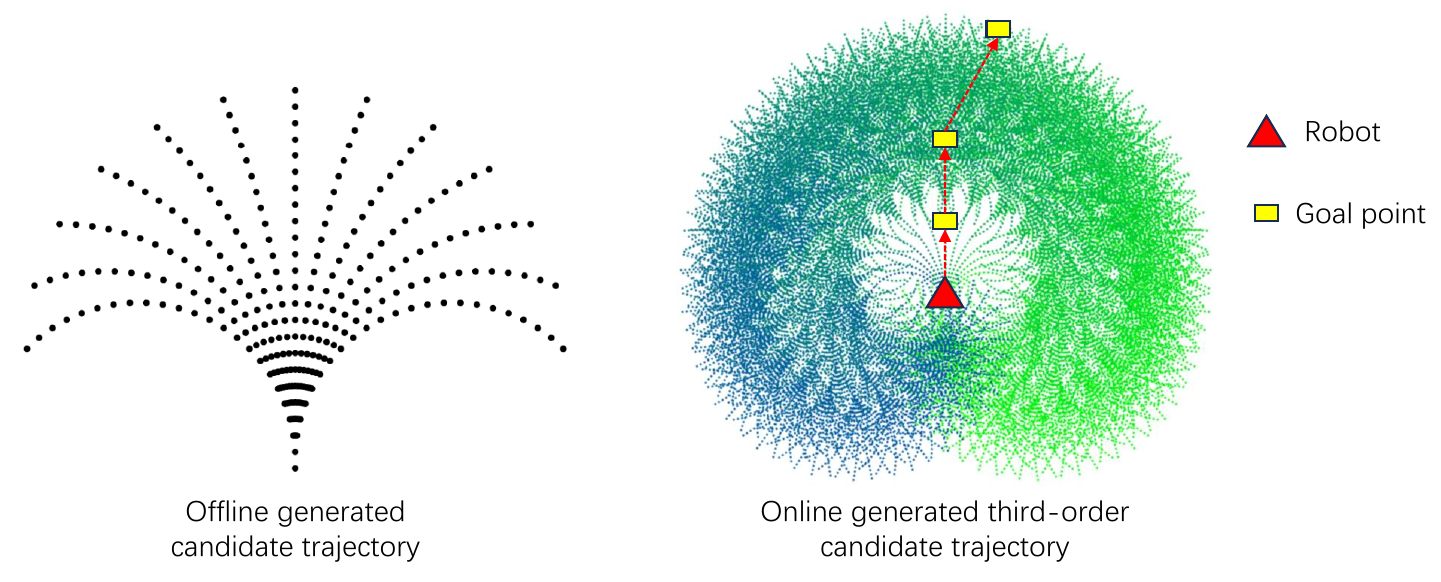}}
	\caption{Our motion planning illustration. We sample several angular velocity values and each corresponds to one candidate trajectory. To increase the covered area, we can repeat the candidate trajectory generation process at each endpoint of each generated trajectory. During the motion planning process, we compute the score of each trajectory at the first group. Angular velocity corresponding to the trajectory with the highest score is sent to the mobile platform.}
	\label{navigation}
\end{figure}

Then, the next goal pose is updated as
\begin{equation}
	\mathbf{T}_{g1}^k = \mathbf{T}_{g0}^k \cdot \mathbf{T}^{S(i)}_{S(S(i))}.
\end{equation}
Thanks to the keyframe list, we are able to compute all the goal poses along the map and formulate a goal list $L_g = \{\mathbf{t}_{g0},\mathbf{t}_{g1},\cdots,\mathbf{t}_{gM}\}$, where $t_g$ is the translational part of $\mathbf{T}_g^k$. The goal list is updated once the robot matches successfully with the map. Notice that the goal poses are always in the frame where the robot matches successfully with the map. Maintaining a goal list can effectively handle the situation of temporarily goal lost due to environmental challenges. Then, our motion module aims to track the goal list. An illustration is presented in Fig. \ref{navigation}.
\subsubsection{Local Motion Planning and Multi-Goal Tracking}
The goal tracking is further developed based on our prior work about local navigation \cite{wang2022lidar}. We construct local trajectory candidates for local motion planning. Specifically, for the robot, its control command vector contains orientation angle and velocity. By fixing velocity, we evenly sample $K$ orientation angles in $[-60^{\circ},60^{\circ}]$. Within a period of 1 second time, we can obtain $K$ trajectory candidates within the current robot frame, which are pre-cached.  The robot is also equipped with a 2D laser sensor in its front part. Given its 2D range observation, we can construct a 2D robot-centric grid map and then we project the local trajectory candidates on the map. After expansion of occupied grids, we directly filter the candidates that pass through occupied grids. For the rest of  candidates, we compute their scores and select the best one, whose corresponding control command vector is then directly published to the robot platform. The trajectory scoring mechanism is detailed as follows. Let $\mathbf{x}$ denote the current position of robot and $\mathbf{t}_e^i$ denote the endpoint of the $i$-th candidate. Then, its score is computed as
\begin{equation}
	s_i = \frac{1}{3}\sum_{m=0}^{m=2}\left(1-(0.005\cdot\Theta(\mathbf{t}_e^i-\mathbf{x},\mathbf{t}_{gm}-\mathbf{x}))^{\frac{1}{2}}\right),
\end{equation}
where $\Theta(\cdot,\cdot)$ returns the angle between two vectors. We consider the first three goals for scoring. If the robot deviates from the route due to obstacle avoidance and $t_{g0}-\mathbf{x}$ is out the scope of $[-60^{\circ},60^{\circ}]$, $t_{g0}$ will be directly erased from $L_g$. If $\mathbf{t}_{gm}$ is determined to be arrived, it will also be erased. The motion planning and control algorithm is presented in Alg. \ref{alg:Framework}. 

\begin{algorithm}[t]
	\caption{Local Motion Planning and Control Algorithm.}
	\label{alg:Framework}
	\begin{algorithmic}[1]    
		\Require
		The environmental obstacle laser point cloud, $P$;
		The relative position coordinates of the goal point list in the robot coordinate system, $x_p, y_p$;
		\Ensure
		Control commands for robot platform: forward linear velocity $v$ and steering angular velocity $\omega $;
		\State $T \gets InitCandidateTrajectory();$
		\State $M_o \gets BuildOccupancyGridMap(P);$
		\State $M \gets MapInflation(M_o);$    
		\State $d_p  \gets Distance(x_p, y_p);$
		\If {$d_p < 0.5$} 
		\If {$Feasibility(M, x_p) =$ True} 
		\State $v = d_p$;
		\State $\omega = atan(y_p / x_p)$;
		\Else
		\State $v = 0$;
		\State $\omega = 0$; 
		\EndIf
		\Else
		\For{$i = 0 \to n - 1$}
		\State $T_{score}[i] = 0$;
		\For{$m = 0 \to k - 1$}
		\If {$Feasibility(M, T[i])  =$ True} 
		\State $x_e, y_e = GetTerminalPoint(T[i])$;
		\State $\theta = \Theta([x_p[m], y_p[m]], [x_e,y_e])$;
		\State $T_{score}[i] += \left(1-(0.005\cdot\theta^{\frac{1}{2}}\right)$;
		\Else
		\State $T_{score}[i] += 0$;
		\EndIf
		\EndFor
		\EndFor
		\State $mI = MaxIndex(T_{score})$
		\State $v = 0.3$;
		\State $\omega = T[mI]$;
		\EndIf    
		\State \Return $v, \omega$
	\end{algorithmic}
\end{algorithm}

\section{Experiments}
\subsection{Experimental Settings}

\subsubsection{Mobile Robot Configuration}
\begin{figure}[t]
	\centerline{\includegraphics[width=0.95\columnwidth]{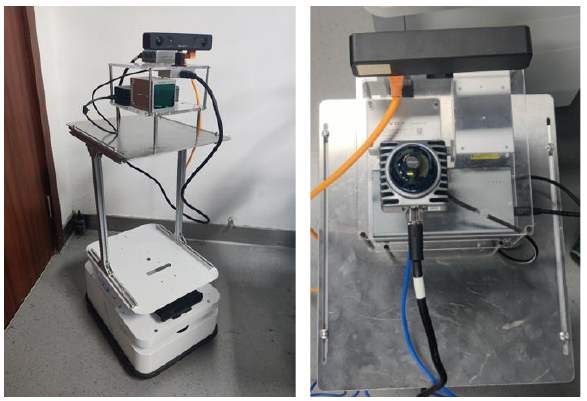}}
	\caption{Our mobile platform. An IMU-embedded stereo camera and 3D LiDAR sensor are mounted for environmental perception and localization. }
	\label{platform}
\end{figure}

Our robot is constructed based on a differential driving platform. A IMU-embedded camera (MYNTEYE-SC) and LiDAR-IMU sersor (Livox Mid-360) are mounted. We implement iG-LIO \cite{10380742} using the LiDAR-IMU sersor observations to record the robot's trajectories for evaluation only. The indoor mobile platform is presented in Fig. \ref{platform}.

\subsubsection{Evaluation}
The widely used evaluation metric is Chamfer distance, which is designed as the geometry difference between the teaching  and   repeating trajectories.  However, the aim of VTR is not to strictly follow the teaching route. In this paper, the VTR is defined as success if the robot can navigate to the end point from the start point, no matter how much deviation from the teaching route during the repeating process. The robot being able to navigating to the end point finally means that the VTR navigation system can effectively correct its actions under the challenges of frame matching aliasing due to environmental changes or obstacle occlusion. Thus, in this paper, the distance between the real arrived end point and the preset end point of teaching route is used as evaluation metric, termed as end-point distance.
\subsubsection{Custom Dataset}
\begin{figure}[t]
	\centerline{\includegraphics[width=0.95\columnwidth]{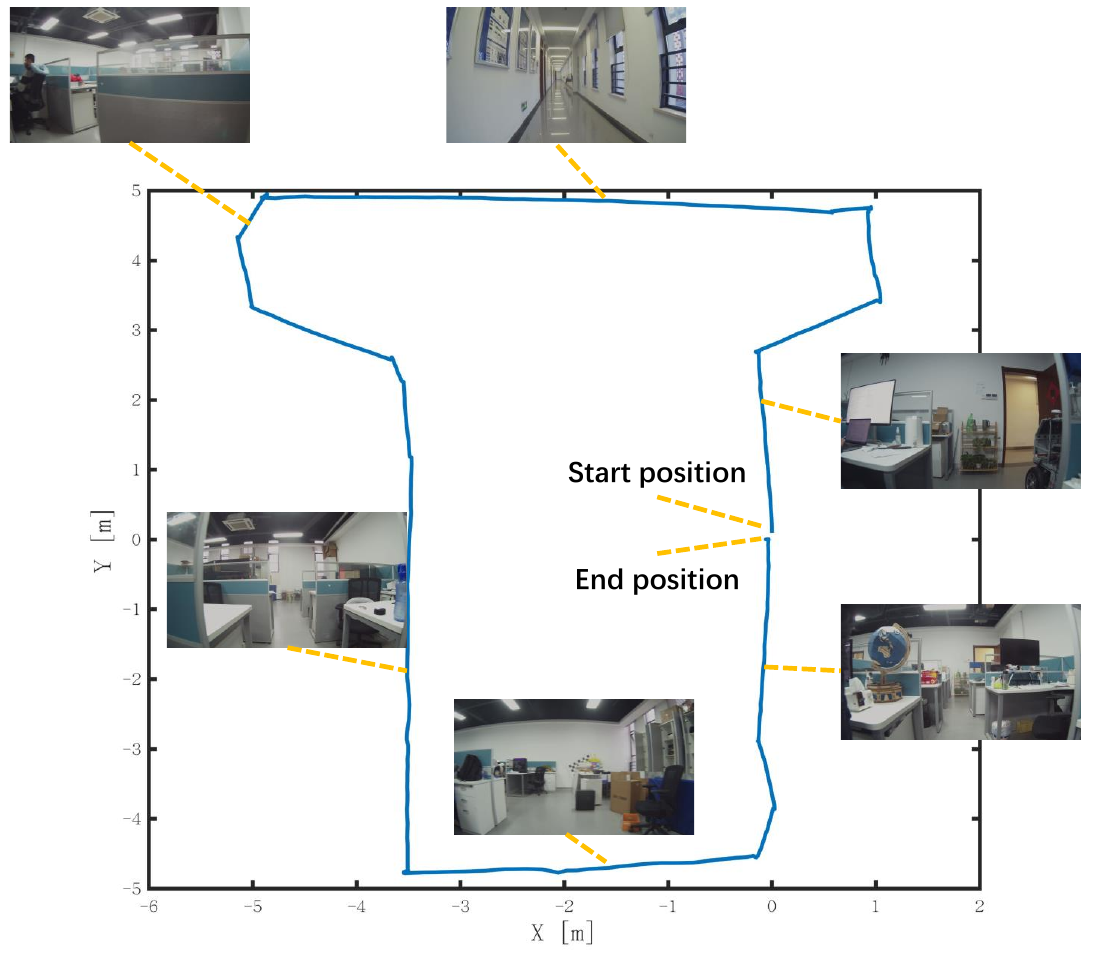}}
	\caption{Typical scenes of our office for teaching route collection. }
	\label{office-teaching}
\end{figure}

We collect dataset in two kinds of scenes, including office and corridor. The teaching trajectory in office is shown in Fig. \ref{office-teaching}. Both straight and curved routes are collected. During the repeating phase, we set two conditions, including normal state, and obstacle occlusion. Under the condition of normal state, the environmental state is nearly same to the teaching phase, and we mainly test the performance of baselines following different kinds of routes. Under condition of obstacle occlusion, the robot will deviate from the route due to obstacle avoidance, and we want to validate that the baselines' ability to restore tracking route.  To validate our method with environmental changing, we implement repeating navigation after teaching phase for several days. Our testing environments are human-living and its appearance changes with time. Appearance changes make it challenged to recognize map keyframe and thus the robustness of baseline's navigation ability can be demonstrated. 

\subsubsection{Baselines}
There exists two kinds of baselines, including bio-inspired methods and visual geometry methods. The bio-inspired methods aims to strictly repeat the teaching route and apply  pattern recognition method to find similarity between repeating and teaching observations. Instead of relying on global descriptors of visual images, the visual geometry methods extract features and make the image matching process more reliable. Both kinds of methods have potential failure once robot deviates from teaching route. Our method can be regarded as a fusion of both kinds of methods. We use DBoW2 for place recognition and use feature points for sophisticated matching. Furthermore, we also propose some robust mechanisms. Thus, we perform thoroughly ablation study.  We also compare our method with the most representative VTR method referring to \cite{dall2021fast}, termed as BVTR\footnote{https://github.com/QVPR/teach-repeat}. BVTR frequently adjusts moving direction according to frame-frame comparison results. In our method, we frequently adjust the goal list to follow the teaching route.

\subsection{Experimental Results}

\subsubsection{Navigation Under Normal Condition}
\begin{figure}[t]
	\centerline{\includegraphics[width=0.95\columnwidth]{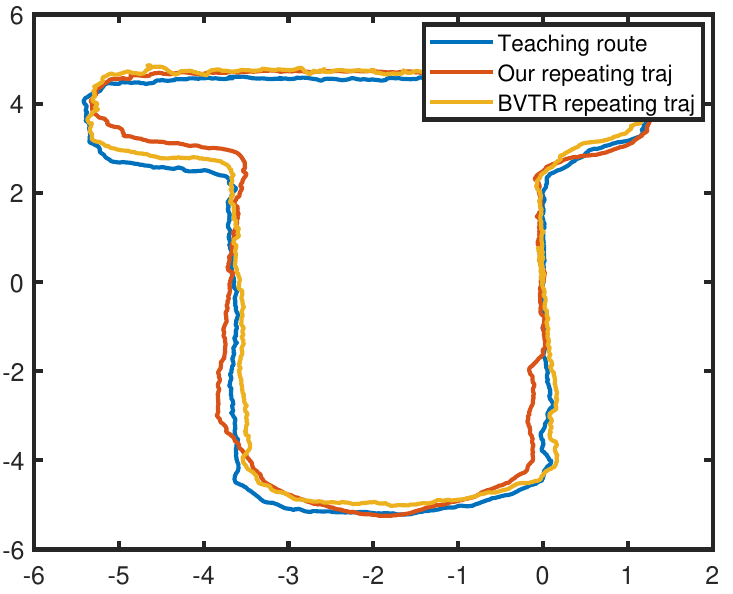}}
	\caption{Repeating navigation results in office scene under normal condition.}
	\label{office-repeating}
\end{figure}

Normal condition means that it is less challenge to match frames with map. We implement repeating navigation in office environment using our method and BVTR. The results are shown in Fig. \ref{office-repeating}. The teaching route contains turning sections and straight road sections.  Both methods can successfully finish repeating navigation. We can tell that little deviation from the teaching route happens at turning sections. However, the robot can effectively correct its trajectory to follow the teaching route at straight road sections. The end-point distances of ours and BVTR are 0.08m and 0.10m, respectively. 

\subsubsection{Navigation with Obstacle on Route}
\begin{figure}[t]
	\centerline{\includegraphics[width=0.95\columnwidth]{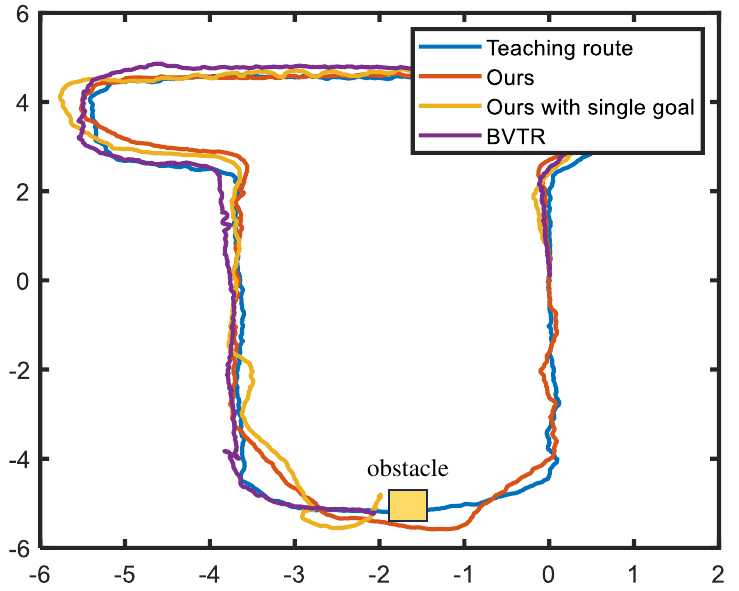}}
	\caption{Repeating navigation results with obstacle on teaching route in office scene.}
	\label{office-obs-repeating}
\end{figure}

To further validate the system's robustness, we put an obstacle on the teaching route. The repeating navigation results are shown in Fig. \ref{office-obs-repeating}. The position of obstacle is shown by a yellow box. With the existence of obstacle, the frame matching module of both methods will lose tracking. We can see that our method can successfully finish the repeating navigation, while BVTR fails. Robot with our system can avoid the obstacle and back to the teaching route. However, robot with BVTR stops at the front of obstacle and do not know what to do since their frame matching module cannot provide effective results to motion planning module. We also modify our method by replacing the proposed multi-goals tracking strategy with one-goal tracking strategy. We can tell that it also fails to avoid the obstacle. It means that our  multi-goals tracking strategy significantly improves the system robustness to obstacle obstruction by its long-term tracking ability. The end-point distances of ours, ours with single goal, and BVTR are 0.08m, 5.20m and 5.58m, respectively.

\subsubsection{Keyframe Cluster Validation}
\label{cluster}
\begin{figure}[t]
	\centerline{\includegraphics[width=0.9\columnwidth]{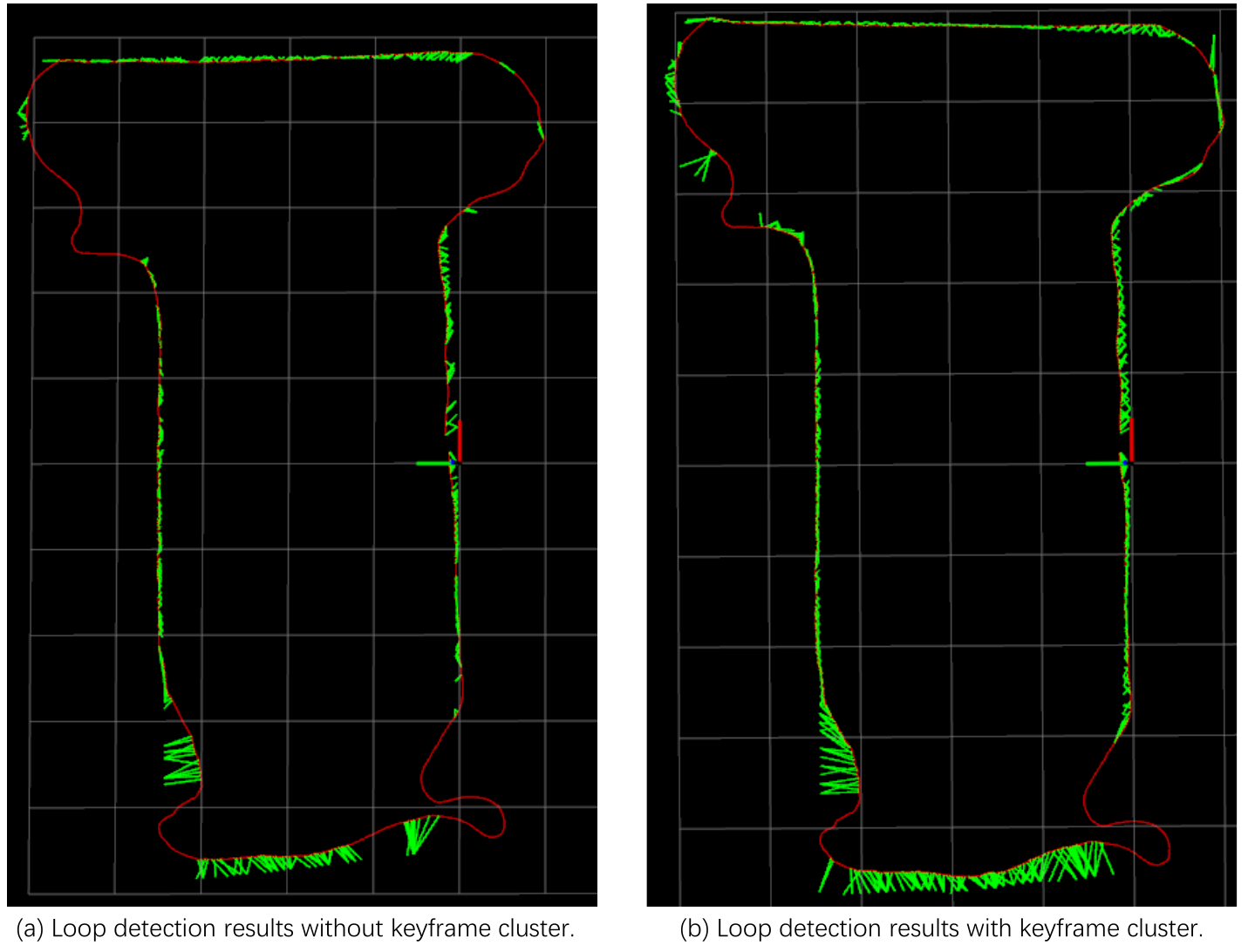}}
	\caption{Loop results without and with keyframe cluster in office scene. Estimated trajectory of the repeating navigation is shown in red lines. The detected loops are shown in blue lines.}
	\label{office-loopresults}
\end{figure}

Frame matching module provides critical cues for repeating navigation. Thus, in this paper, we cluster keyframes to increase the number of the observed feature points on one keyframe. The success rate of frame matching is positively correlated to the involved feature points' scale. With the constructed keyframe map using the teaching route, we control the robot to loosely follow the teaching route and implement loop detection in real-time. The results are shown in Fig. \ref{office-loopresults}. Green lines indicate detected loops. The test trajectory is shown in red lines. We can tell that loop detection results based on our keyframe clustering strategy are more dense then that of raw keyframe map, especially at turning sections. Dense loop detection can timely provide feedback cues for motion planning module and thus promote the visual repeating performance. 

\subsubsection{Map Expansion Validation}

\begin{figure}[t]
	\centerline{\includegraphics[width=0.9\columnwidth]{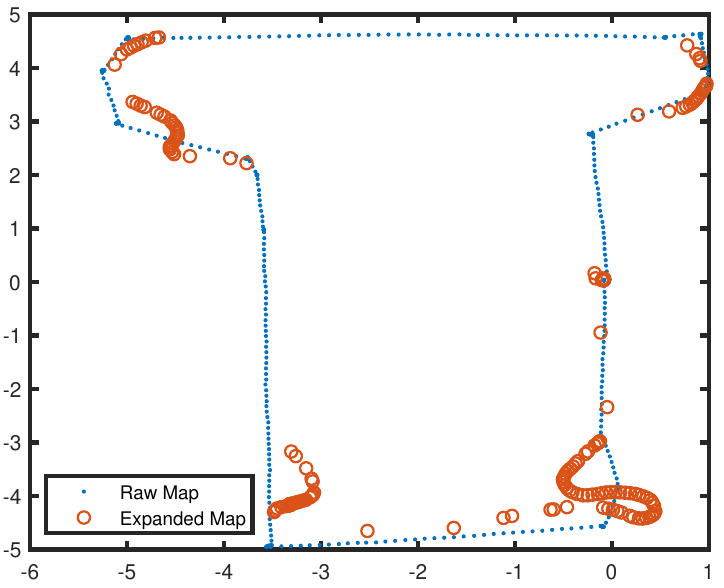}}
	\caption{Map expansion result. Keyframes in raw map are shown in blue dots. The expanded novel keyframes are shonw in orange circles. These novel keyframes are that fail to formulate loops with the raw map.}
	\label{office-mapexpansion}
	\vspace{-0cm}
\end{figure}

During the navigation process, the environment changes and the robot may explore new area. It is a critical issue that the robot starts repeating navigation from unknown place. Thus, the map should have the ability to be expanded to fuse novel environmental information without lose its topological structure. To validate the map expansion strategy, we test trajectory in Section. \ref{cluster} is used, since it is not strictly following the teaching route and novel environmental information is involved. We implement map expansion and the results are shown in Fig. \ref{office-mapexpansion}. We can see that novel sections are added to the expanded map. Each added keyframe are associated with one keyframe in the raw map. These keyframes fails to find loops with the raw map thus are extracted and added to the raw map. Notice that each expanded keyframe has a pointer pointed to its closed keyframe in the raw map. 

\subsubsection{Navigation on Curved Teaching Route}
\begin{figure}[t]
	\centerline{\includegraphics[width=0.95\columnwidth]{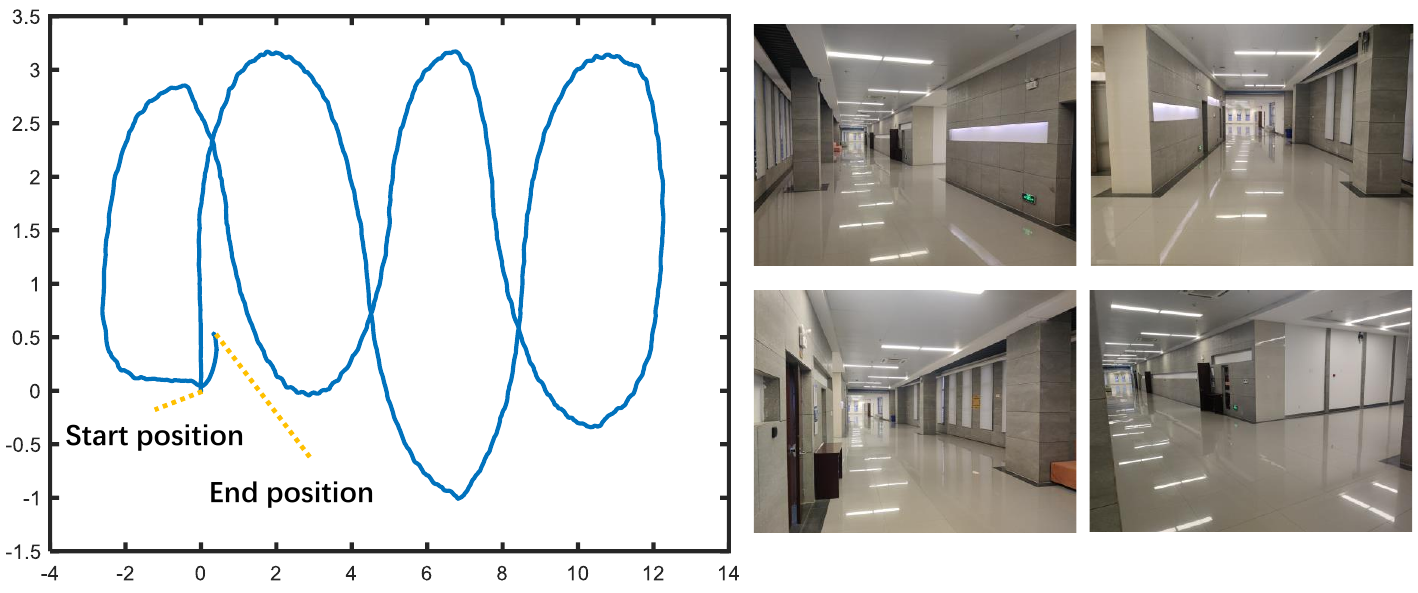}}
	\caption{Teaching route in corridor scene, which demonstrates texture-less.}
	\label{corridor-teach}
\end{figure}

\begin{figure}[t]
	\centerline{\includegraphics[width=0.95\columnwidth]{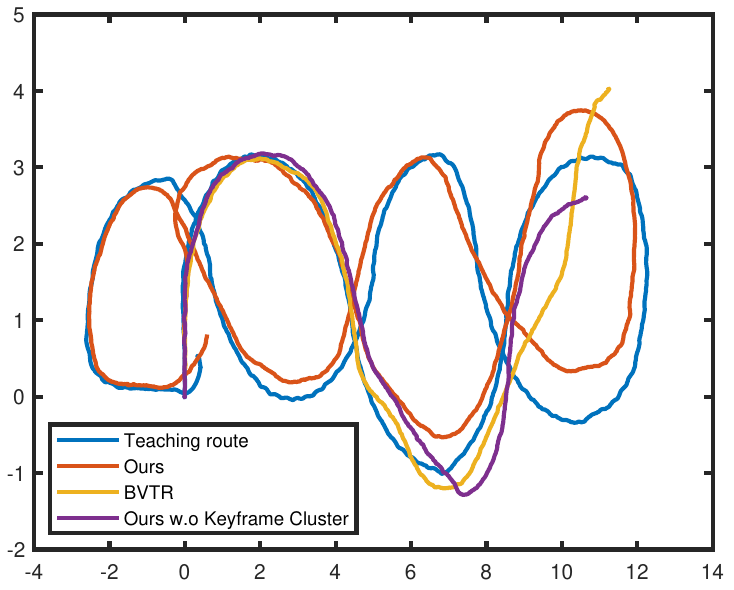}}
	\caption{Repeating navigation results in corridor scene under normal condition.}
	\label{comp1-raw}
\end{figure}

\begin{figure}[t]
	\centerline{\includegraphics[width=0.95\columnwidth]{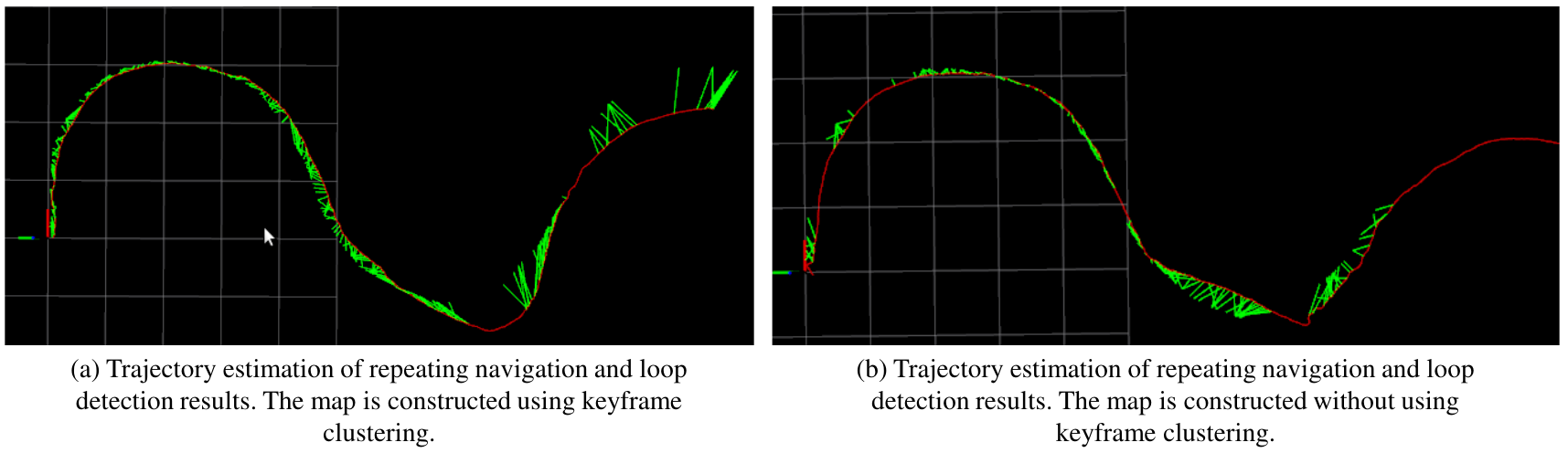}}
	\caption{Estimated trajectory of the repeating navigation using our method without keyframe clustering. In (a), we offline estimate the trajectory and find loops with our map using keyframe clustering. In (b), we offline estimate the trajectory and find loops with our map without using keyframe clustering. More loops are detected in (a).}
	\label{loopcomp}
\end{figure}

\begin{figure}[t]
	\centerline{\includegraphics[width=0.95\columnwidth]{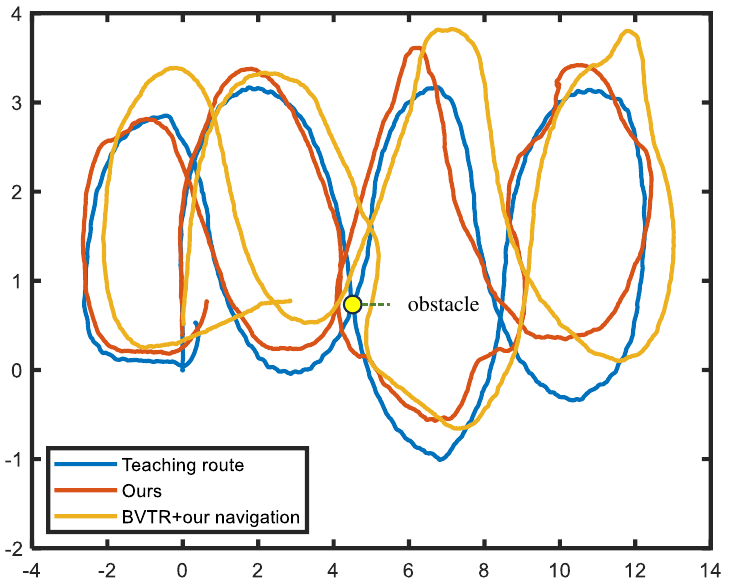}}
	\caption{Repeating navigation results in corridor scene with obstacle on teaching route.}
	\label{comp2-obs}
\end{figure}

\begin{figure}[t]
	\centerline{\includegraphics[width=0.95\columnwidth]{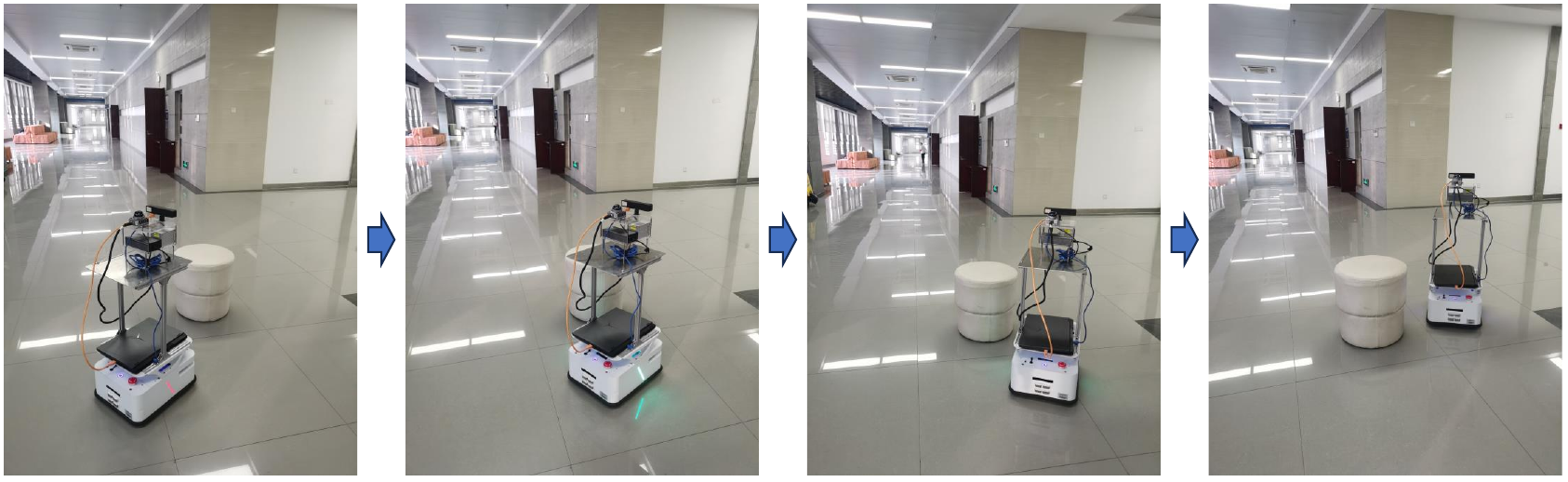}}
	\caption{Obstacle avoidance process of our method.}
	\label{corridor-obsimg}
\end{figure}

To further validate our system, we implement VTR experiments in corridor environment, which is texture-less and exhibits visual aliasing.  The teaching route is also highly curved, which can validate the route following ability. The teaching route and scene are shown in Fig. \ref{corridor-teach}.  Firstly, we perform directly visual repeating navigation and the repeating trajectories of the methods are shown in Fig. \ref{comp1-raw}. The end-point distances of ours and BVTR are 0.37m and 11.44m, respectively. We can tell that our method can successfully follow the teaching route, while BVTR fails. When testing BVTR, the robot navigates to an unknown place and stops without further action due to the fail of frame matching. Though our method can follow the entire teaching route, the route deviation is larger than that on straight line sections. Furthermore, we also perform ablation study of keyframe clustering of our method. Our method without keyframe clustering is tested and the result is also demonstrated in Fig. \ref{comp1-raw}. Its corresponding end-point distance is 10.49, which illustrate that keyframe clustering is critical for robust frame matching in texture-less environment. We also offline validate the loop detection results of its trajectory. The results are shown in Fig. \ref{loopcomp}. We can tell that more loops are detected with our map using keyframe clustering. At the end of the trajectory, no loop is found, thus ours without keyframe clustering fails to repeat the left route. Secondly, we test the methods' robustness to obstacle occlusion. Since BVTR fails in normal condition, we replace its navigation module with our navigation module. The repeating trajectories are shown in Fig. \ref{comp2-obs}. The end-point distances of ours and modified BVTR are 0.36m and 2.51m, respectively. Both methods successfully follow the entire route and escape from the obstacle. Even though our navigation module effectively improves BVTR's performance, its route following quality is also limited. We can tell that its real end position deviates significantly from the preset end position.  Obstacle avoidance process of our method is shown in Fig. \ref{corridor-obsimg}.  

\subsection{Discussion}
The core of visual teach-and-repeat navigation is the alleviation of geometric map construction and real-time global pose tracking. It is achieved by the combination of loop detection and local navigation.  The highlights of our proposed method are summarized as follows. 

Firstly, a novel topo-metric map representation is proposed, which demonstrates highly flexibility. It can preserve the main topology of the teaching route and support novel environmental information insertion, which is important for robots implementing long-term navigation tasks in changing environment. Furthermore, our map is also supportive for teaching route with graph structure, which can realize multiple point-to-point VTR navigation. 

Secondly, in this paper, loop detection performance is promoted by keyframe clustering. Loop detection performance play critical role for VTR navigation. However, the existences of visual aliasing and environmental changes are unavoidable, which tend to make loop detection fail. Though deep learning-based place recognition methods have been developed, highly computational cost is required. Some simple frame matching methods are also proposed before, which are directly matching the raw image data. Such mechanism is based on the assumption that the robot is always on the teaching route without deviation, which is impractical. Ours loop detection is based on visual features, which is robust to viewpoint changing. Our strategy can also promote the  storage consumption while improving the loop detection performance. 

Thirdly, a long-term goal tracking strategy  is proposed, which makes our system robust to loop detection fail or route deviation. Experimental results demonstrate that long-term goal tracking can help robot escape from being trapped by obstacle or loop detection fail, compared with single goal tracking. The effectiveness of the long-term goal management heavily depends on the truth that the relative poses between keyframes are reliable. Since the reliability decreases with distance, we update the long-term goal list once the robot get new loop detection. Essentially, the long-term goal tracking ability make the system adaptable to low-frequency frame-map matching. However, due to the odometry drift errors, if the robot cannot find matches with map in a long time, its deviation from teaching route  tends to diverge, which is hard to be corrected. An illustration is the failed repeating navigation of our method without keyframe clustering in corridor scene.

\section{Conclusion}
This paper proposes a novel visual teach-and-repeat navigation method, which can help the robot to be deployed in task environment efficiently. The robot collects environmental information on the human-guided teaching route and formulate a novel flexible topo-metric map. Based on the map, the robot can follow the teaching route robustly with our frame matching and motion planning module. Our system is adaptable to obstacle occlusion, complex teaching route, environmental changing, and task environment extending. In the future, we will construct a deep-learning-based model to realize a more intelligent end-to-end visual navigation, which also gets rid of the requirement of accurately global pose tracking and environmental mapping. 

\begin{IEEEbiography}[{\includegraphics[width=1in,height=1.25in,clip,keepaspectratio]{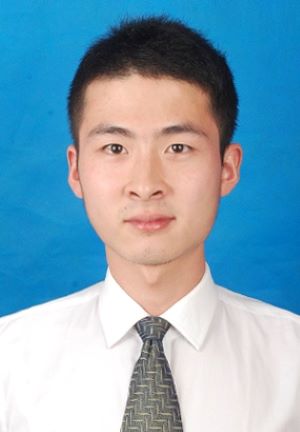}}]{JiKai Wang} was born Anhui, China, in 1993. He received the B.S. degree in electrical engineering and automation from the University of Yanshan, Qinhuangdao, China, in 2014,and the Ph.D. degree in control science and engineering from the University of Science and Technology (USTC) of China, Hefei, China, in2020.
	He is currently a Special Associate Researcher with the Department of Automation,USTC. His research interests include 3-D dense mapping based on Gaussian splatting, multi-sensor fusion-based SLAM systems, intelligent robotic and information processing, and mobile robot visual teach-and-repeat in GNSS-denied environments.
	
	Dr. Wang is the Secretary of the Simulation Technology Application Special Committee of the China Simulation Federation. He is also a Member of the System Simulation Committee of the China Association of Automation.
	\end{IEEEbiography}

	\begin{IEEEbiography}[{\includegraphics[width=1in,height=1.25in,clip,keepaspectratio]{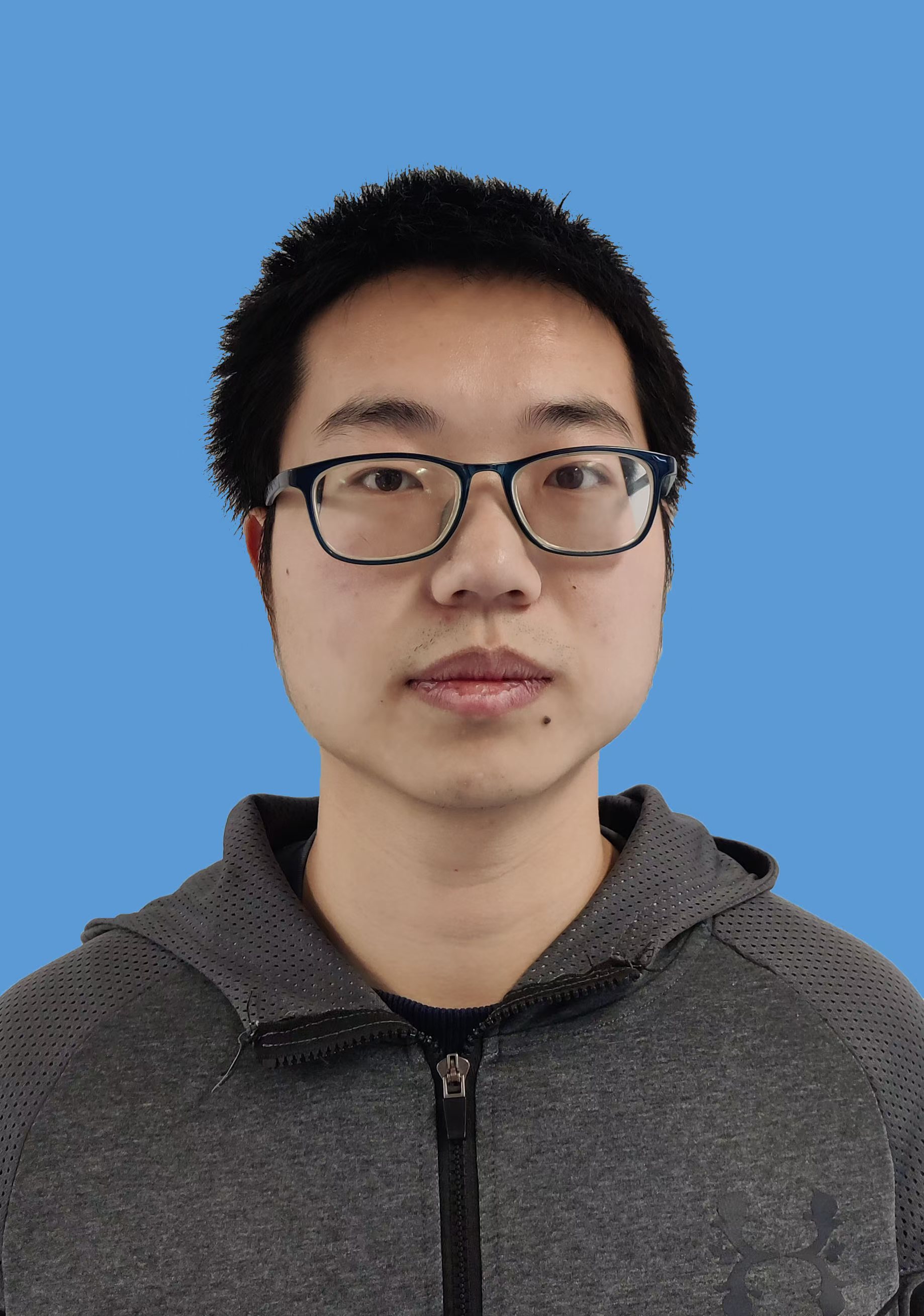}}]{Yunqi Cheng}  was born in Shaanxi, China, in 2001. He received the B.S. degree from Nankai University, in 2023. He is now a master degree candidate in control science and engineering from the Department of Automation, University of Science and Technology of China, Hefei, China. His research interests include simultaneous localization and mapping (SLAM) and mobile robot navigation.
		\end{IEEEbiography}
		
		\begin{IEEEbiography}[{\includegraphics[width=1in,height=1.25in,clip,keepaspectratio]{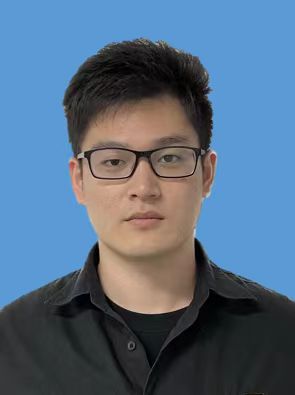}}]{Kezhi Wang}  was born in Shandong, China, in 1998. He received the B.S. degree in automation from Chongqing University, Chongqing,China, in 2021. He is currently working toward the Ph.D. degree in control science and engineering from the Department of Automation,University of Science and Technology of China,Hefei, China.
	
			His research interests include simultaneous localization and mapping (SLAM), 3-D reconstruction, robotics, and multi-sensor fusion.
		\end{IEEEbiography}
		
		\begin{IEEEbiography}[{\includegraphics[width=1in,height=1.25in,clip,keepaspectratio]{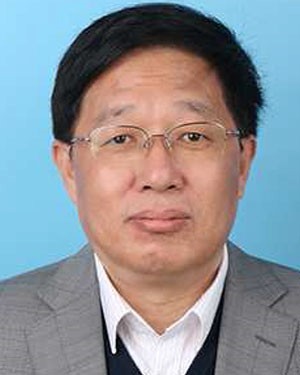}}]{Zonghai Chen} (Senior Member, IEEE) was born in Anhui, China, in 1963. He received the B.S. degree in automation and the M.E. degree in control theory and control engineering from the University of Science and Technology of China (USTC), Hefei, China, in 1988 and 1991, respectively.
			
			He has been a Professor with the Department of Automation, USTC, since 1998. His research interests include modeling and control of complex systems, intelligent robotic and information
			processing, energy management technologies for electric vehicles, andsmart microgrids.
			
			Prof. Chen was a recipient of special allowances from the State Council of PR China. He is a Member of the Robotics Technical Committeeand Modelling, Identification, and Signal Processing Technical Committee of the International Federation of Automation Control (IFAC). 
		\end{IEEEbiography}

\vfill


\begin{thebibliography}{00}
	
	\bibitem{10036102}Wang, X., Fu, H., Deng, G., Liu, C., Tang, K. \& Chen, C. Hierarchical Free Gait Motion Planning for Hexapod Robots Using Deep Reinforcement Learning. {\em IEEE Transactions On Industrial Informatics}. \textbf{19}, 10901-10912 (2023)
	
	\bibitem{10521763}Yin, J., Zhuang, Y., Yan, F., Liu, Y. \& Zhang, H. A Tightly-Coupled and Keyframe-Based Visual-Inertial-Lidar Odometry System for UGVs With Adaptive Sensor Reliability Evaluation. {\em IEEE Transactions On Systems, Man, And Cybernetics: Systems}. \textbf{54}, 4976-4985 (2024)
	
	\bibitem{8931657}Yuan, J., Zhu, W., Dong, X., Sun, F., Zhang, X., Sun, Q. \& Huang, Y. A Novel Approach to Image-Sequence-Based Mobile Robot Place Recognition. {\em IEEE Transactions On Systems, Man, And Cybernetics: Systems}. \textbf{51}, 5377-5391 (2021)
	
	\bibitem{9316980}Wang, R., Zhang, X., Fang, Y. \& Li, B. Virtual-Goal-Guided RRT for Visual Servoing of Mobile Robots With FOV Constraint. {\em IEEE Transactions On Systems, Man, And Cybernetics: Systems}. \textbf{52}, 2073-2083 (2022)
	
	\bibitem{van2024visual}Dijk, T., De Wagter, C. \& Croon, G. Visual route following for tiny autonomous robots. {\em Science Robotics}. \textbf{9}, eadk0310 (2024)
	
	\bibitem{10433735}Ghafourian, A., CuiZhu, Z., Shi, D., Chuang, I., Charette, F., Sachdeva, R. \& Soltani, I. Hierarchical End-to-End Autonomous Navigation Through Few-Shot Waypoint Detection. {\em IEEE Robotics And Automation Letters}. \textbf{9}, 3211-3218 (2024)
	
	\bibitem{sun2021robust}Sun, L., Taher, M., Wild, C., Zhao, C., Zhang, Y., Majer, F., Yan, Z., Krajniik, T., Prescott, T. \& Duckett, T. Robust and long-term monocular teach and repeat navigation using a single-experience map. {\em 2021 IEEE/RSJ International Conference On Intelligent Robots And Systems (IROS)}. pp. 2635-2642 (2021)
	
	\bibitem{dall2021fast}Dall’Osto, D., Fischer, T. \& Milford, M. Fast and robust bio-inspired teach and repeat navigation. {\em 2021 IEEE/RSJ International Conference On Intelligent Robots And Systems (IROS)}. pp. 500-507 (2021)
	
	\bibitem{furgale2010visual}Furgale, P. \& Barfoot, T. Visual teach and repeat for long-range rover autonomy. {\em Journal Of Field Robotics}. \textbf{27}, 534-560 (2010)
	
	\bibitem{paton2016bridging}Paton, M., MacTavish, K., Warren, M. \& Barfoot, T. Bridging the appearance gap: Multi-experience localization for long-term visual teach and repeat. {\em 2016 IEEE/RSJ International Conference On Intelligent Robots And Systems (IROS)}. pp. 1918-1925 (2016)
	
	\bibitem{zheng2013revisiting}Zheng, Y., Kuang, Y., Sugimoto, S., Astrom, K. \& Okutomi, M. Revisiting the pnp problem: A fast, general and optimal solution. {\em Proceedings Of The IEEE International Conference On Computer Vision}. pp. 2344-2351 (2013)
	
	\bibitem{swedish2018deep}Swedish, T. \& Raskar, R. Deep visual teach and repeat on path networks. {\em Proceedings Of The IEEE Conference On Computer Vision And Pattern Recognition Workshops}. pp. 1533-1542 (2018)
	
	\bibitem{gao2017intention}Gao, W., Hsu, D., Lee, W., Shen, S. \& Subramanian, K. Intention-net: Integrating planning and deep learning for goal-directed autonomous navigation. {\em Conference On Robot Learning}. pp. 185-194 (2017)
	
	\bibitem{ai2022deep}Ai, B., Gao, W., Hsu, D. \& Others Deep visual navigation under partial observability. {\em 2022 International Conference On Robotics And Automation (ICRA)}. pp. 9439-9446 (2022)
	
	\bibitem{sorokin2022learning}Sorokin, M., Tan, J., Liu, C. \& Ha, S. Learning to navigate sidewalks in outdoor environments. {\em IEEE Robotics And Automation Letters}. \textbf{7}, 3906-3913 (2022)
	
	\bibitem{ullah2024mobile}Ullah, I., Adhikari, D., Khan, H., Anwar, M., Ahmad, S. \& Bai, X. Mobile robot localization: Current challenges and future prospective. {\em Computer Science Review}. \textbf{53} pp. 100651 (2024)
	
	
	
	
	\bibitem{zhou2020ego}Zhou, X., Wang, Z., Ye, H., Xu, C. \& Gao, F. Ego-planner: An esdf-free gradient-based local planner for quadrotors. {\em IEEE Robotics And Automation Letters}. \textbf{6}, 478-485 (2020)
	
	\bibitem{loo2024scene}Loo, J. \& Hsu, D. Scene Action Maps: Behavioural Maps for Navigation without Metric Information. {\em ArXiv Preprint ArXiv:2405.07948}. (2024)
	
	\bibitem{cuizhu2023one}CuiZhu, Z., Charette, F., Ghafourian, A., Shi, D., Cui, M., Krishnamachar, A. \& Soltani, I. One-Shot Learning of Visual Path Navigation for Autonomous Vehicles. {\em ArXiv Preprint ArXiv:2306.08865}. (2023)
	
	\bibitem{garg2024robohop}Garg, S., Rana, K., Hosseinzadeh, M., Mares, L., Sünderhauf, N., Dayoub, F. \& Reid, I. Robohop: Segment-based topological map representation for open-world visual navigation. {\em ArXiv Preprint ArXiv:2405.05792}. (2024)
	
	\bibitem{paul2024mpvo}Paul, S., Bhowmick, B. \& Others MPVO: Motion-Prior based Visual Odometry for PointGoal Navigation. {\em ArXiv Preprint ArXiv:2411.04796}. (2024)
	
	\bibitem{10578334}Rouček, T., Rozsypálek, Z., Blaha, J., Ulrich, J. \& Krajník, T. Predictive Data Acquisition for Lifelong Visual Teach, Repeat and Learn. {\em IEEE Robotics And Automation Letters}. \textbf{9}, 10042-10049 (2024)
	
	\bibitem{rozsypalek2023multidimensional}Rozsypálek, Z., Roucek, T., Vintr, T. \& Krajnik, T. Multidimensional particle filter for long-term visual teach and repeat in changing environments. {\em IEEE Robotics And Automation Letters}. \textbf{8}, 1951-1958 (2023)
	
	\bibitem{krajnik2018navigation}Krajnik, T., Majer, F., Halodova, L. \& Vintr, T. Navigation without localisation: reliable teach and repeat based on the convergence theorem. {\em 2018 IEEE/RSJ International Conference On Intelligent Robots And Systems (IROS)}. pp. 1657-1664 (2018)

	\bibitem{geneva2020openvins}Geneva, P., Eckenhoff, K., Lee, W., Yang, Y. \& Huang, G. Openvins: A research platform for visual-inertial estimation. {\em 2020 IEEE International Conference On Robotics And Automation (ICRA)}. pp. 4666-4672 (2020)
	
	\bibitem{9440682}Campos, C., Elvira, R., Rodríguez, J., M. Montiel, J. \& D. Tardós, J. ORB-SLAM3: An Accurate Open-Source Library for Visual, Visual–Inertial, and Multimap SLAM. {\em IEEE Transactions On Robotics}. \textbf{37}, 1874-1890 (2021)
	
	\bibitem{qin2017vins}Qin, T., Li, P. \& Shen, S. VINS-Mono: A Robust and Versatile Monocular Visual-Inertial State Estimator. {\em IEEE Transactions On Robotics}. \textbf{34}, 1004-1020 (2018)
	
	\bibitem{9484819}Lin, S., Wang, J., Xu, M., Zhao, H. \& Chen, Z. Topology Aware Object-Level Semantic Mapping Towards More Robust Loop Closure. {\em IEEE Robotics And Automation Letters}. \textbf{6}, 7041-7048 (2021)
	
	\bibitem{10042487}Lin, S., Wang, J., Xu, M., Zhao, H. \& Chen, Z. Contour-SLAM: A Robust Object-Level SLAM Based on Contour Alignment. {\em IEEE Transactions On Instrumentation And Measurement}. \textbf{72} pp. 1-12 (2023)
	
	\bibitem{wang2022lidar}Wang, K., Li, J., Xu, M., Chen, Z. \& Wang, J. LiDAR-Only Ground Vehicle Navigation System in Park Environment. {\em World Electric Vehicle Journal}. \textbf{13}, 201-220 (2022)
	
	\bibitem{10380742}Chen, Z., Xu, Y., Yuan, S. \& Xie, L. iG-LIO: An Incremental GICP-Based Tightly-Coupled LiDAR-Inertial Odometry. {\em IEEE Robotics And Automation Letters}. \textbf{9}, 1883-1890 (2024)
	
\end{thebibliography}
\end{document}